\documentclass{article}

\usepackage[final,nonatbib]{tackling_climate_workshop_style}

\usepackage[utf8]{inputenc} 
\usepackage[T1]{fontenc}    
\usepackage{hyperref}       
\usepackage{url}            
\usepackage{booktabs}       
\usepackage{amsfonts}       
\usepackage{nicefrac}       
\usepackage{microtype}      
\usepackage{xcolor}         
\usepackage{amsmath, amsfonts, graphicx, booktabs, tabularx, multirow, siunitx, xcolor, soul, bm, enumitem, tikz}

\DeclareMathOperator*{\argmin}{arg\,min}

\graphicspath{{./figures/}}

\title{Addressing Deep Learning Model Uncertainty in Long-Range Climate Forecasting with Late Fusion}

\author{%
    \phantom{aaaa}Ken C. L. Wong\thanks{Corresponding author (\texttt{clwong@us.ibm.com})}\phantom{aaaa}\\
    \phantom{aaaa}IBM Research\phantom{aaaa}\\
    \phantom{aaaa}San Jose, CA, USA\phantom{aaaa}\\
    \And
    \phantom{aaaa}Hongzhi Wang\phantom{aaaa}\\
    \phantom{aaaa}IBM Research\phantom{aaaa}\\
    \phantom{aaaa}San Jose, CA, USA\phantom{aaaa}\\
    \And
    Etienne E. Vos\\
    IBM Research\\
    Johannesburg, GP, South Africa\\
    \And
    Bianca Zadrozny\\
    IBM Research\\
    Rio De Janeiro, RJ, Brazil\\
    \And
    Campbell D. Watson\\
    IBM Research\\
    Yorktown Heights, NY, USA\\
    \And
    Tanveer Syeda-Mahmood\\
    IBM Research\\
    San Jose, CA, USA\\
}

\begin{document}

\maketitle

\begin{abstract}
  Global warming leads to the increase in frequency and intensity of climate extremes that cause tremendous loss of lives and property. Accurate long-range climate prediction allows more time for preparation and disaster risk management for such extreme events. Although machine learning approaches have shown promising results in long-range climate forecasting, the associated model uncertainties may reduce their reliability. To address this issue, we propose a late fusion approach that systematically combines the predictions from multiple models to reduce the expected errors of the fused results. We also propose a network architecture with the novel denormalization layer to gain the benefits of data normalization without actually normalizing the data. The experimental results on long-range 2m temperature forecasting show that the framework outperforms the 30-year climate normals, and the accuracy can be improved by increasing the number of models.
\end{abstract}

\section{Introduction}

Global warming leads to the increase in frequency and intensity of climate extremes \cite{Journal:Lenton:Nature2019:climate}. High-impact extreme events such as heat waves, cold fronts, floods, droughts, and tropical cyclones can result in tremendous loss of lives and property, and accurate predictions of such events benefit multiple sectors including water, energy, health, agriculture, and disaster risk reduction \cite{Journal:Merryfield:BAMS2020:current}. The longer the range of an accurate prediction, the more the time for proper preparation and response. Therefore, accurate long-range forecasting of the key climate variables such as precipitation and temperature is valuable.

Numerical models for weather and climate prediction have a long history of producing the most accurate seasonal and multi-annual climate forecasts, but they come with the cost of large and expensive physics-based simulations (e.g. \cite{Journal:Johnson:GMD2019:seas5,Journal:Doi:AMES2016:improved}). With the recent advancements in machine learning such as deep learning, the use of machine learning for climate forecasting has become more popular \cite{Journal:Ham:Nature2019:deep,Journal:Yen:SR2019:application,Workshop:Vos:NIPSWorkshop2020,Workshop:Rodrigues:ICMLWorkshop2021}, and some machine learning approaches can outperform numerical models in certain tasks \cite{Journal:Ham:Nature2019:deep}. Nevertheless, depending on the machine learning algorithm and data availability, different degrees of model uncertainties exist. In deep learning, models trained with the same data and hyperparameters are usually not identical. This is caused by the random processes in training such as weight initialization and data shuffling. Such model uncertainties can be more prominent for climate forecasting given the limited data, and this can reduce the reliability of the models especially with large lead times. Even though reducing the randomness in training (e.g., using fixed weight initialization) may reduce the model uncertainties, the chance of getting better models is also reduced.

In this paper, our goal is to reduce model uncertainties and improve accuracy in seasonal climate forecasting. By modifying the late fusion approach in \cite{wang2021modeling} to adapt to deep learning regression, predictions from different models trained with identical hyperparameters are systematically combined to reduce the expected errors in the fused results. We demonstrate its applicability on long-range 2m temperature forecasting. Furthermore, we propose a novel denormalization layer which allows us to gain the benefits of data normalization without actually normalizing the data.

\section{Methodology}


\subsection{Network Architecture with Denormalization}

\begin{figure}[t]
    \centering
    \includegraphics[width=1\linewidth]{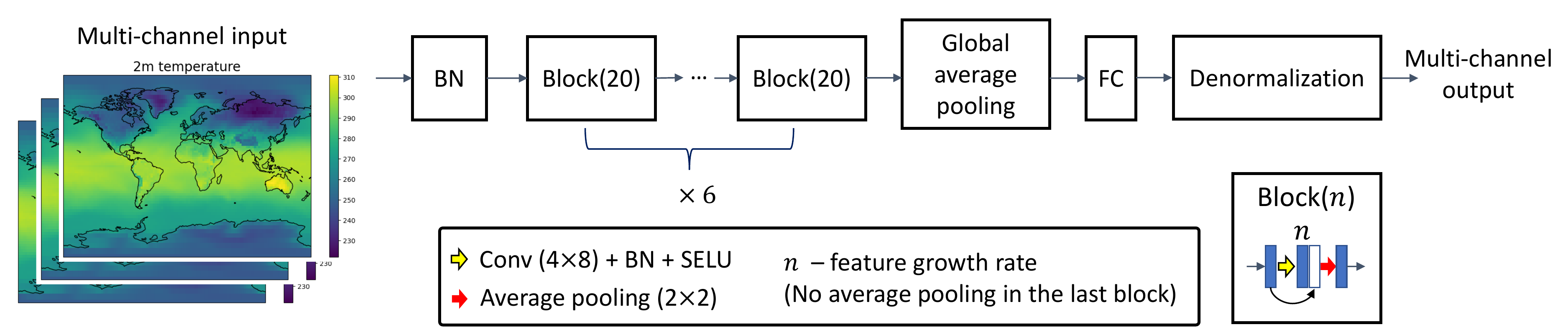}
    \caption{Network architecture for 2m temperature forecasting. BN and FC represent batch normalization and fully connected layers, respectively. The number of input channels is the input horizon, and the number of output channels is the number of predictand locations.}
    \label{fig:network}
\end{figure}

The proposed network architecture is shown in Fig. \ref{fig:network}. Given a multi-channel input tensor formed by stacking the input maps of 2m temperature spanning a fixed input horizon, the network predicts the 2m temperatures at multiple locations with a fixed lead time. The network comprises six dense blocks \cite{Conference:Huang:CVPR2017}, each with a convolutional layer and a growth rate of 20. A batch normalization layer is used right after the input layer for data normalization. Furthermore, although we found that normalizing the predictands allows the use of simpler architectures with better accuracy, the resulting model can only provide normalized predictions and postprocessing is required to recover the original values. To address this issue, we introduce a \emph{denormalization} layer after the final fully connected layer to obtain:
\begin{equation}\label{eq:denormalization}
  x_o(c) = x_i(c) \sigma(c) + m(c)
\end{equation}
with $c$ the channel index, and $x_o$, $x_i$ the output and input features, respectively. $\sigma$ and $m$ are the standard deviation and mean value computed from the training data. Using this denormalization layer, the final fully connected layer only needs to predict normalized values, thus removing the need of predictand normalization. With this architecture, data normalization in training and forecast denormalization in inference are unnecessary.

\subsection{Late Fusion}
We modified the late fusion approach in \cite{wang2021modeling} for regression. The method combines predictions from multiple models using weighted average. To compute the weights, the pairwise correlations between different models in terms of how likely they will make correlated errors are estimated, which are then used to compute the weights that reduce the expected error in the fused result. Let $f^j(s_i)$ be the prediction by the $j_{th}$ model for input $s_i$ and $t(s_i)$ the true value. The late fusion result for $s_i$ is $\sum_j w^j f^j(s_i)$ with $\sum_j w^j=1$. The pairwise correlation between model $j_1$ and $j_2$ is:
\begin{equation}\label{e:M}
  M[j_1,j_2]=\sum_i \left[f^{j_1}(s_i)-t(s_i)\right]\left[f^{j_2}(s_i)-t(s_i)\right]
\end{equation}
Then the weights are computed by:
\begin{equation}\label{e:w}
  \mathbf{w} = \argmin_{\mathbf{w}} \mathbf{w}^\mathrm{T} \mathbf{M} \mathbf{w} = \frac{\mathbf{M}^{-1}\mathbf{1}_K}{\mathbf{1}_K^\mathrm{T} \mathbf{M}^{-1}\mathbf{1}_K}
\end{equation}
with $K$ the number of models and $\mathbf{1}_K$ a vector with $K$ ones. $\mathbf{M}$ and $\mathbf{w}$ are computed using the validation data. This procedure is applied on each output channel.

\subsection{Training Strategy}

The 2m temperature maps of the ERA5 reanalysis data \cite{Journal:Hersbach:JRMS2020:era5} were partitioned for training (1979 -- 2007), validation (2008 -- 2011), and testing (2012 -- 2020). Each data map was resampled from the original spatial resolution of $0.25^{\circ}\times0.25^{\circ}$ to $1^{\circ}\times1^{\circ}$. The data were also aggregated over time from hourly to weekly. An input horizon of six weeks was used with 10 forecast lead times (5 to 50 weeks with a stride of 5 weeks). Each model was trained for 200 epochs with the batch size of 32. The Nadam optimizer \cite{Workshop:Dozat:2016:Nadam} was used with the cosine annealing learning rate scheduler \cite{Conference:Loshchilov:ICLR2017:SGDR}, with the minimum and maximum learning rates as $10^{-4}$ and $10^{-2}$, respectively. The mean absolute error was used as the loss function.

\begin{table}[t]
\caption{Locations at low or high latitudes where the 2m temperatures are predicted.}
\label{table:locations}
\scriptsize
\centering
\begin{tabular*}{\columnwidth}{@{\extracolsep{\fill}}ll}
\toprule
\textbf{Low} & Honolulu (21.3$^{\circ}$N, 157.9$^{\circ}$W), Panama City (9.0$^{\circ}$N, 79.5$^{\circ}$W), Singapore (1.4$^{\circ}$N, 103.8$^{\circ}$E), Mid Paciﬁc Ocean (4.4$^{\circ}$N, 167.7$^{\circ}$W) \\
\midrule
\textbf{High} & Moscow (55.8$^{\circ}$N, 37.6$^{\circ}$E), London (51.5$^{\circ}$N, 0.1$^{\circ}$W), Christchurch (43.5$^{\circ}$S, 172.6$^{\circ}$E), Perth (32.0$^{\circ}$S,115.9$^{\circ}$E) \\
\bottomrule
\end{tabular*}
\end{table}

\begin{figure}[t]
    \centering
    \begin{minipage}[t]{0.34\linewidth}
        \scriptsize
        \centering
        \begin{tikzpicture}
            \node[anchor=south west,inner sep=0] (image) at (0,0) {\includegraphics[width=1\linewidth]{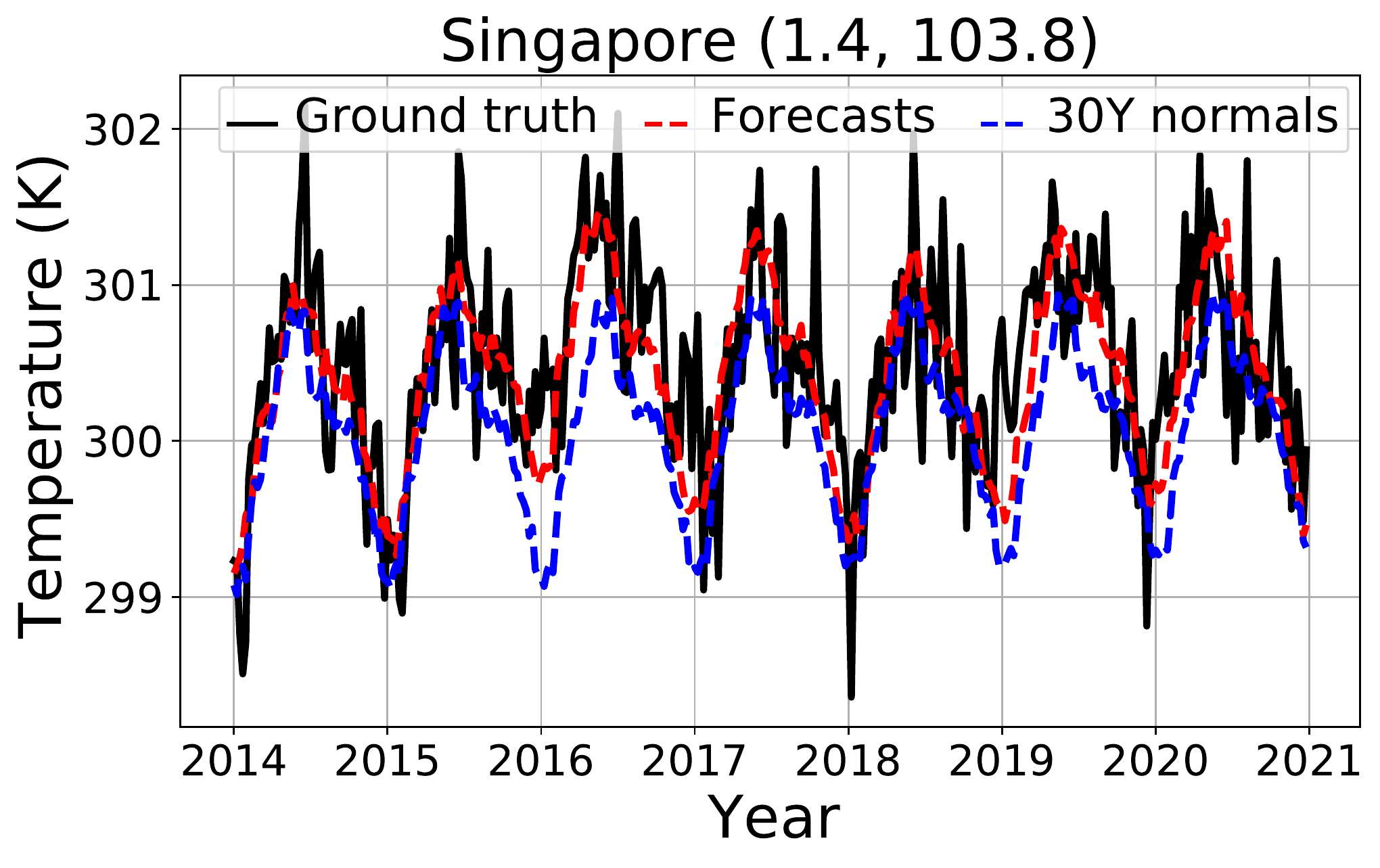}};
            \begin{scope}[x={(image.south east)},y={(image.north west)}]
                \draw[green,thick] (0.42,0.53) ellipse (1.2em and 3em);
            \end{scope}
        \end{tikzpicture}
    \end{minipage}
    \begin{minipage}[t]{0.34\linewidth}
        \scriptsize
        \centering
        \includegraphics[width=1\linewidth]{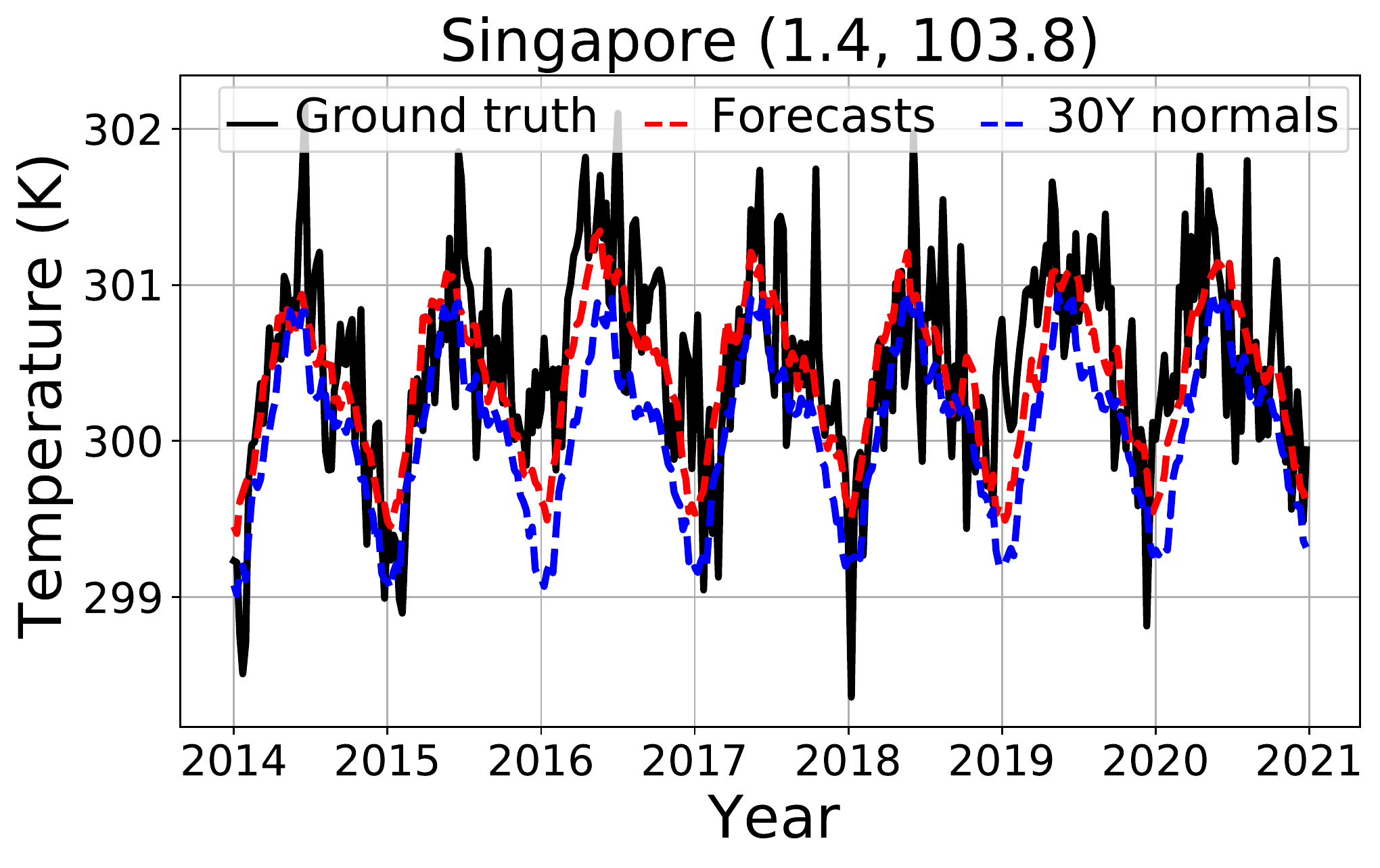}
    \end{minipage}
    \begin{minipage}[t]{0.30\linewidth}
        \scriptsize
        \centering
        \includegraphics[width=1\linewidth]{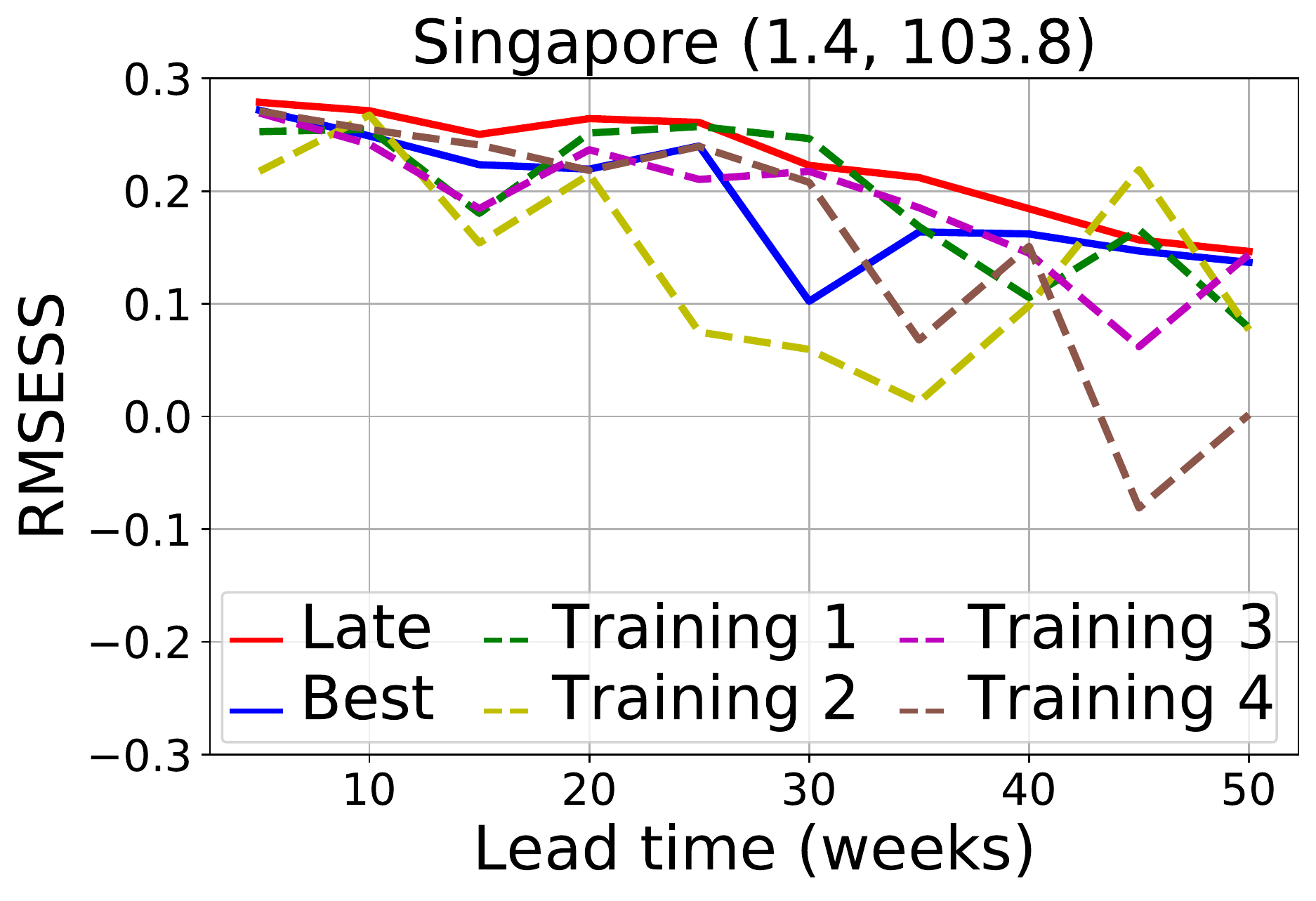}
    \end{minipage}
    \\
    \begin{minipage}[t]{0.34\linewidth}
        \scriptsize
        \centering
        \includegraphics[width=1\linewidth]{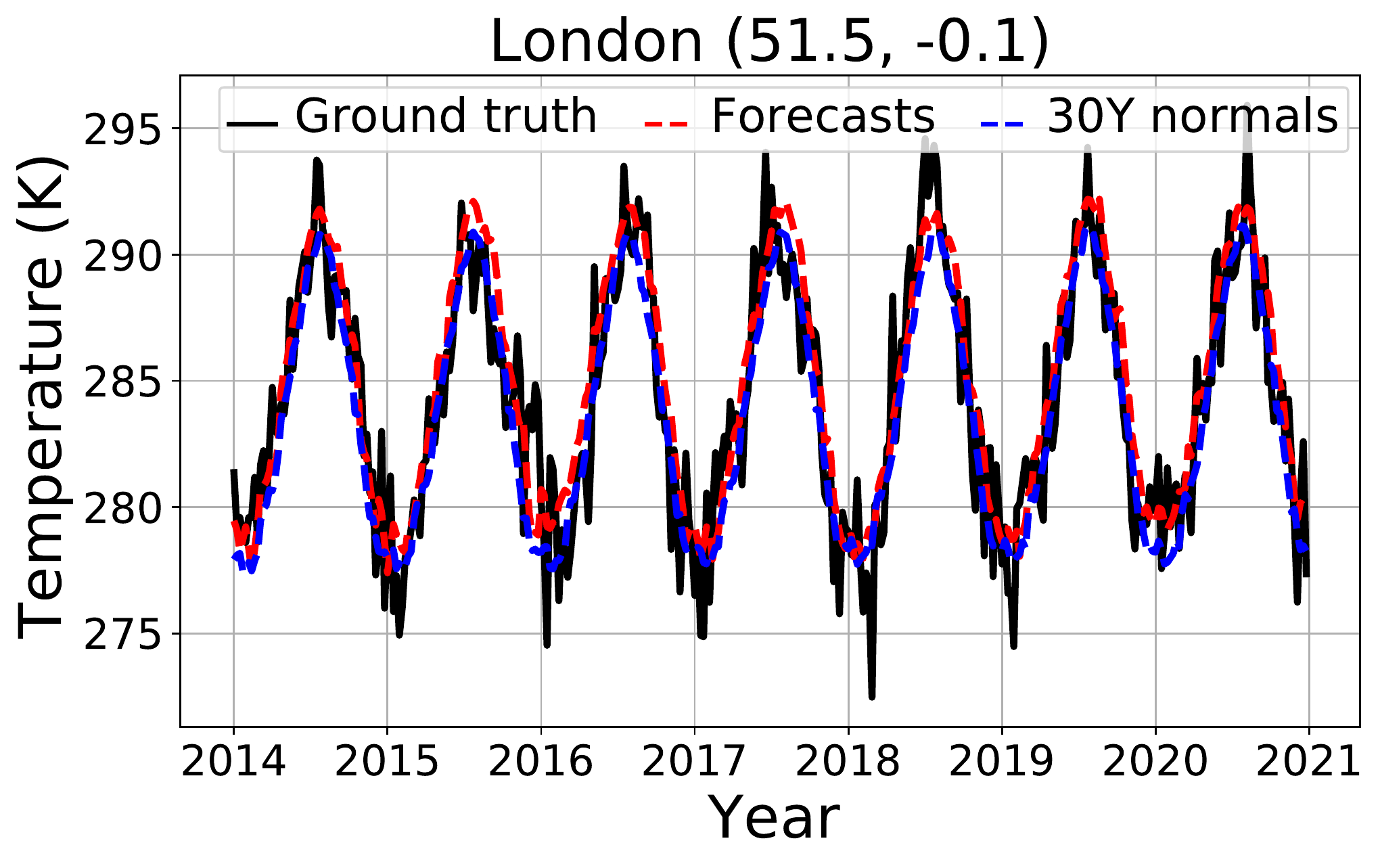}
        \centering{\phantom{aaaa}\textbf{Lead time = 5 weeks}}
    \end{minipage}
    \begin{minipage}[t]{0.34\linewidth}
        \scriptsize
        \centering
        \includegraphics[width=1\linewidth]{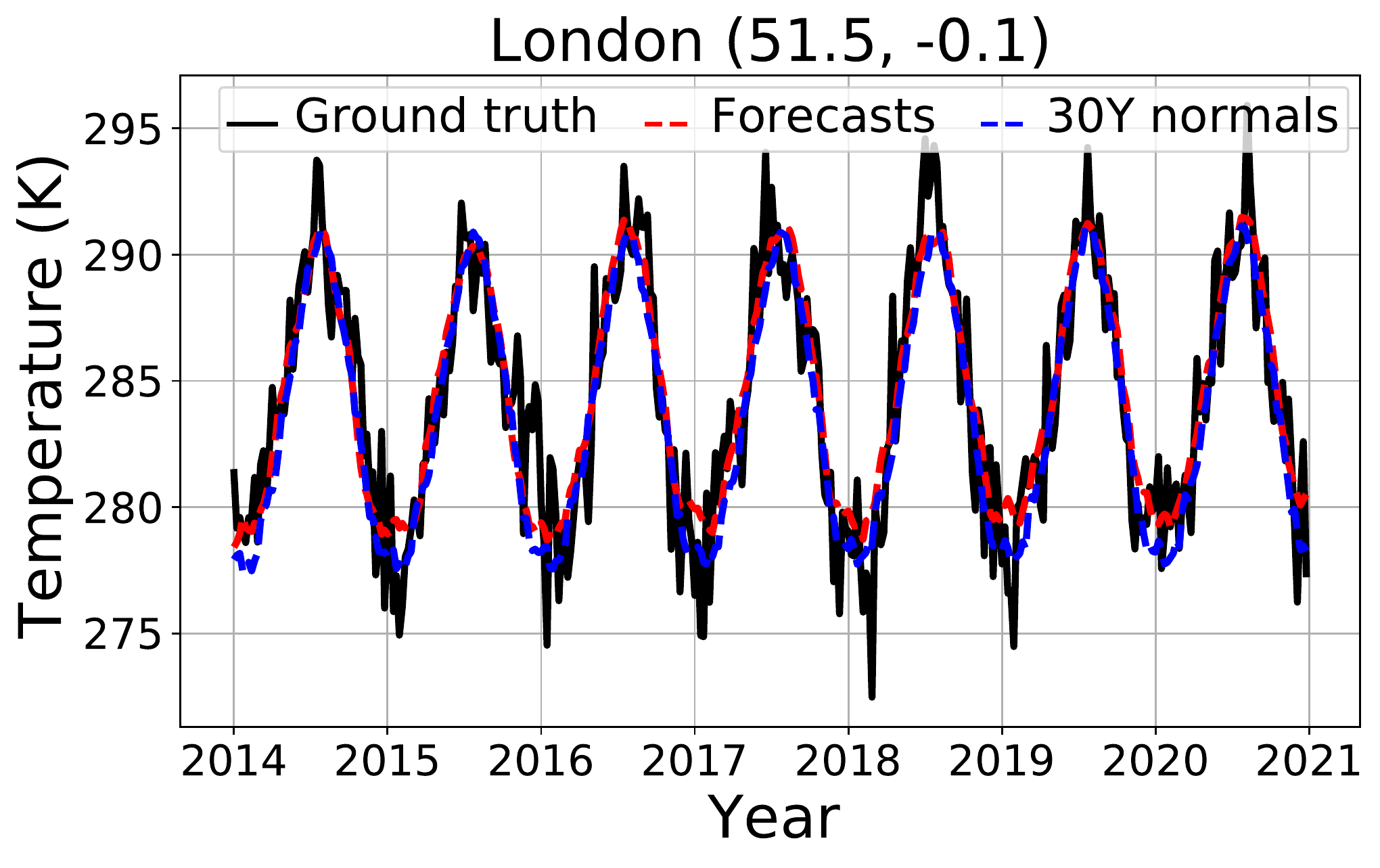}
        \centering{\phantom{aaaa}\textbf{Lead time = 30 weeks}}
    \end{minipage}
    \begin{minipage}[t]{0.30\linewidth}
        \scriptsize
        \centering
        \includegraphics[width=1\linewidth]{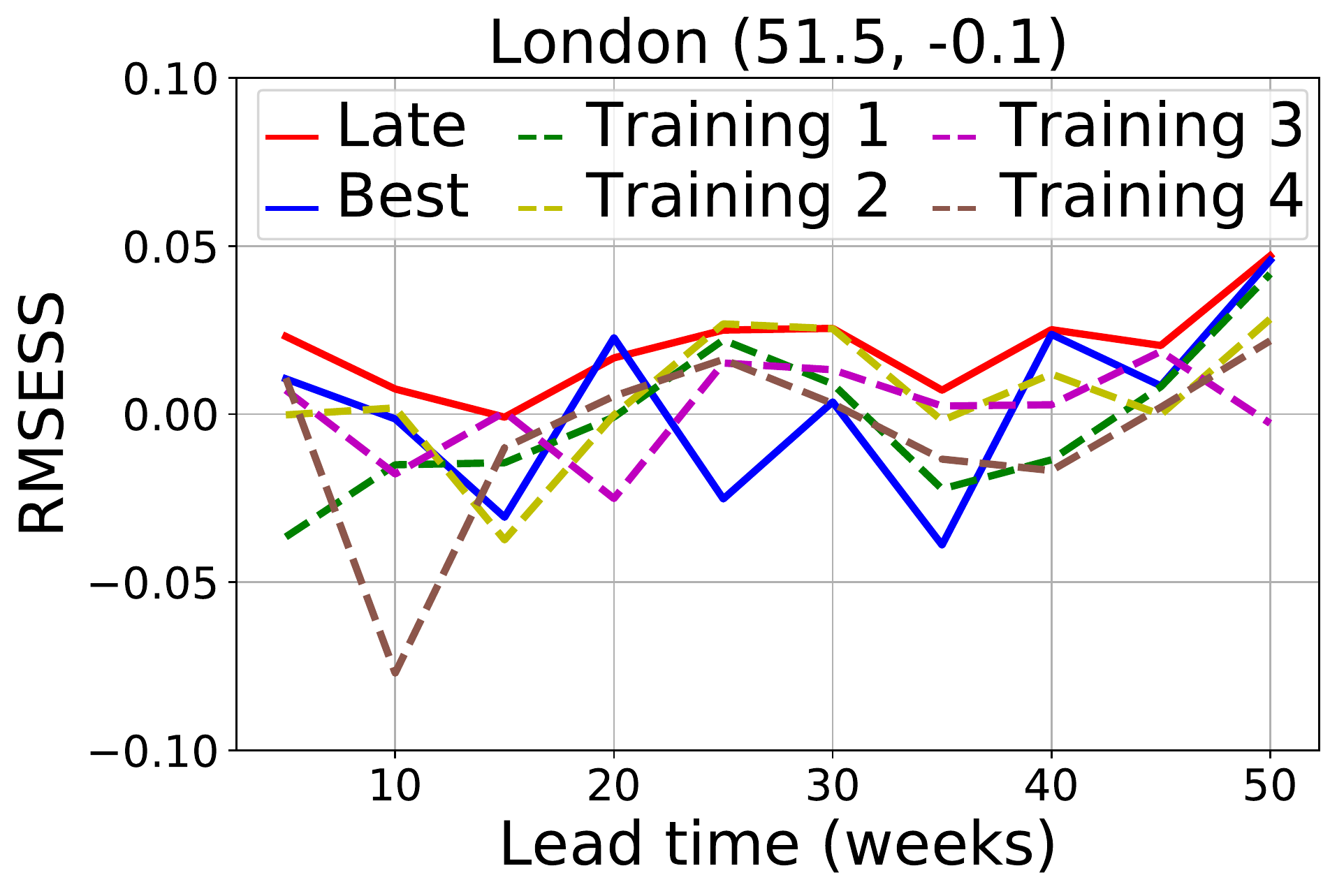}
        \centering{\phantom{aaaa}\textbf{RMSESS vs. lead time}}
    \end{minipage}
    \smallskip
    \caption{Left two: examples of forecasts at different lead times, with the green circle highlighting the hottest year (2016) on record. Right: RMSESS of models trained with identical hyperparameters (dashed lines) compared with the late fusion and the best model frameworks with 20 models per lead time (solid lines). Top: Singapore. Bottom: London.}
    \label{fig:predictions}
\end{figure}

\section{Experiments}

To study model uncertainties, we trained 20 models with identical hyperparameters per lead time (i.e., 200 models in total). Each model was used to predict temperatures from four low-latitude and four high-latitude locations (Table \ref{table:locations}). Two frameworks were compared:
\begin{itemize}[leftmargin=1em]
  \item \textbf{Late fusion}: the framework that combines the predictions of different models at each lead time.
  \item \textbf{Best model}: at each lead time, the model with the smallest root mean square error (RMSE) on the validation data was chosen to provide the predictions.
\end{itemize}
For evaluation, the RMSE skill score (RMSESS $\in [-\infty, 1]$) that compares between the model forecasts and the 30-year climate normals was used:
\begin{equation}
\label{eq:rmsess}
    \mathrm{RMSESS} = 1 - \frac{\mathrm{RMSE_{model}}}{\mathrm{RMSE_{clim}}}
\end{equation}
with $\mathrm{RMSE_{model}}$ computed between the forecasts and the true values, and $\mathrm{RMSE_{clim}}$ computed between the 30-year climate normals and the true values. A 30-year climate normal is the 30-year average of a predictand at a given time point, which is a generally accepted benchmark for comparison.

\begin{figure}[t]
    \centering
    \begin{minipage}[t]{0.40\linewidth}
        \scriptsize
        \centering
        \includegraphics[width=1\linewidth]{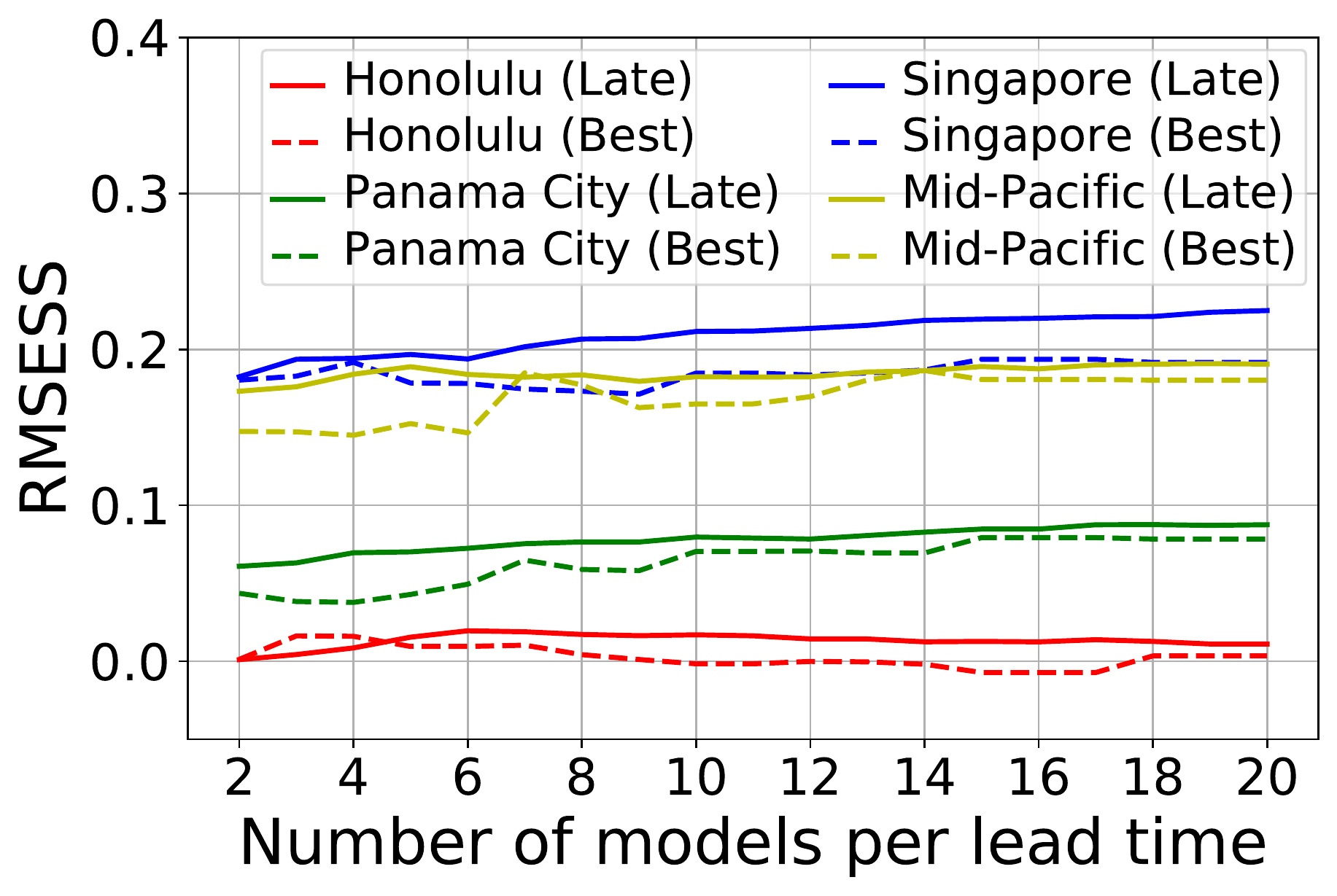}
    \end{minipage}
    \hspace{2em}
    \begin{minipage}[t]{0.42\linewidth}
        \scriptsize
        \centering
        \includegraphics[width=1\linewidth]{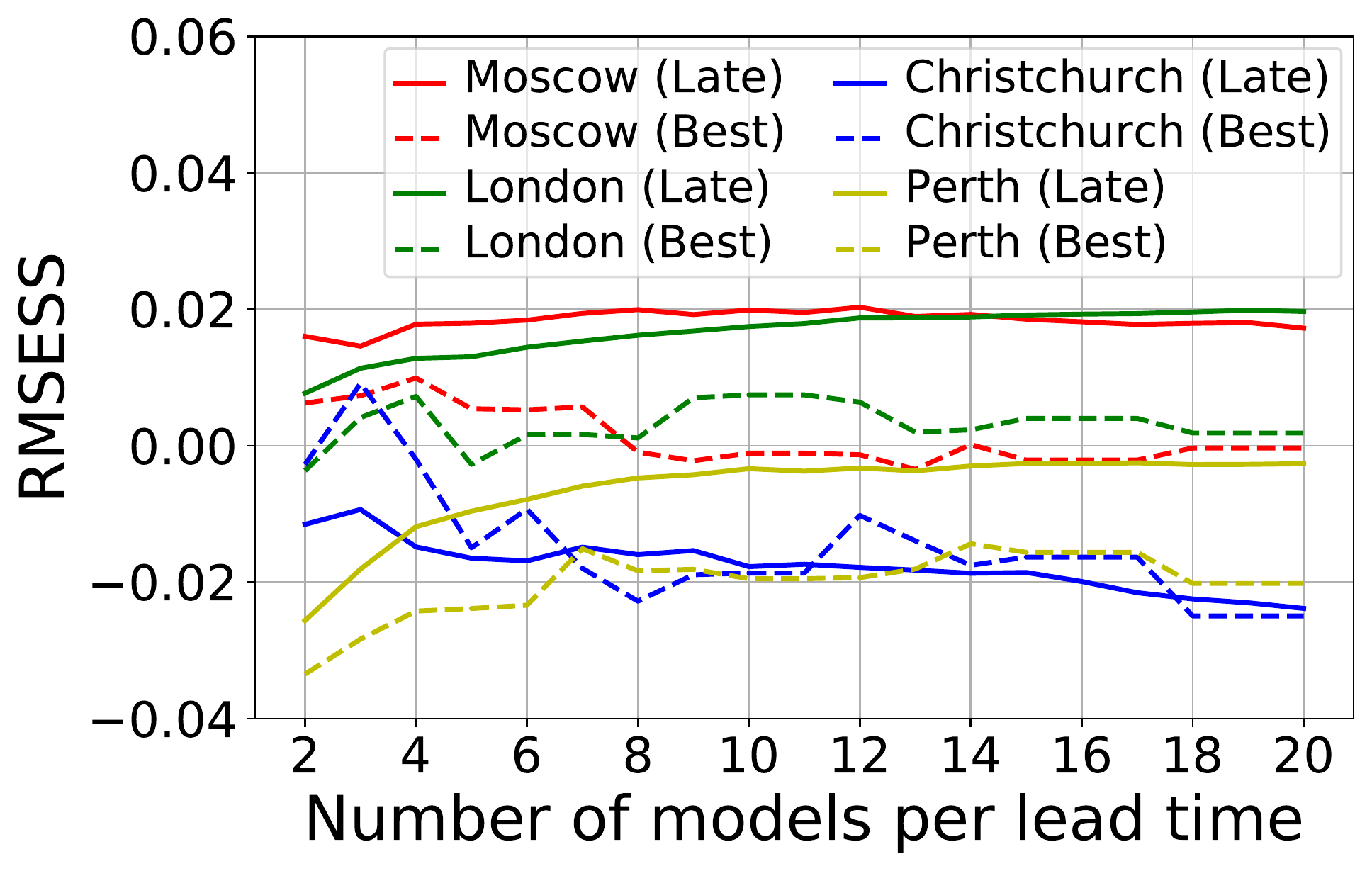}
    \end{minipage}
    \caption{Comparison between the late fusion and the best model frameworks. The y-axis shows the average RMSESS over the lead times. Left to right: low-latitude and high-latitude locations.}
    \label{fig:vs_num_models}
\end{figure}

\begin{figure}[t]
    \centering
    \begin{minipage}[t]{0.40\linewidth}
        \scriptsize
        \centering
        \includegraphics[width=1\linewidth]{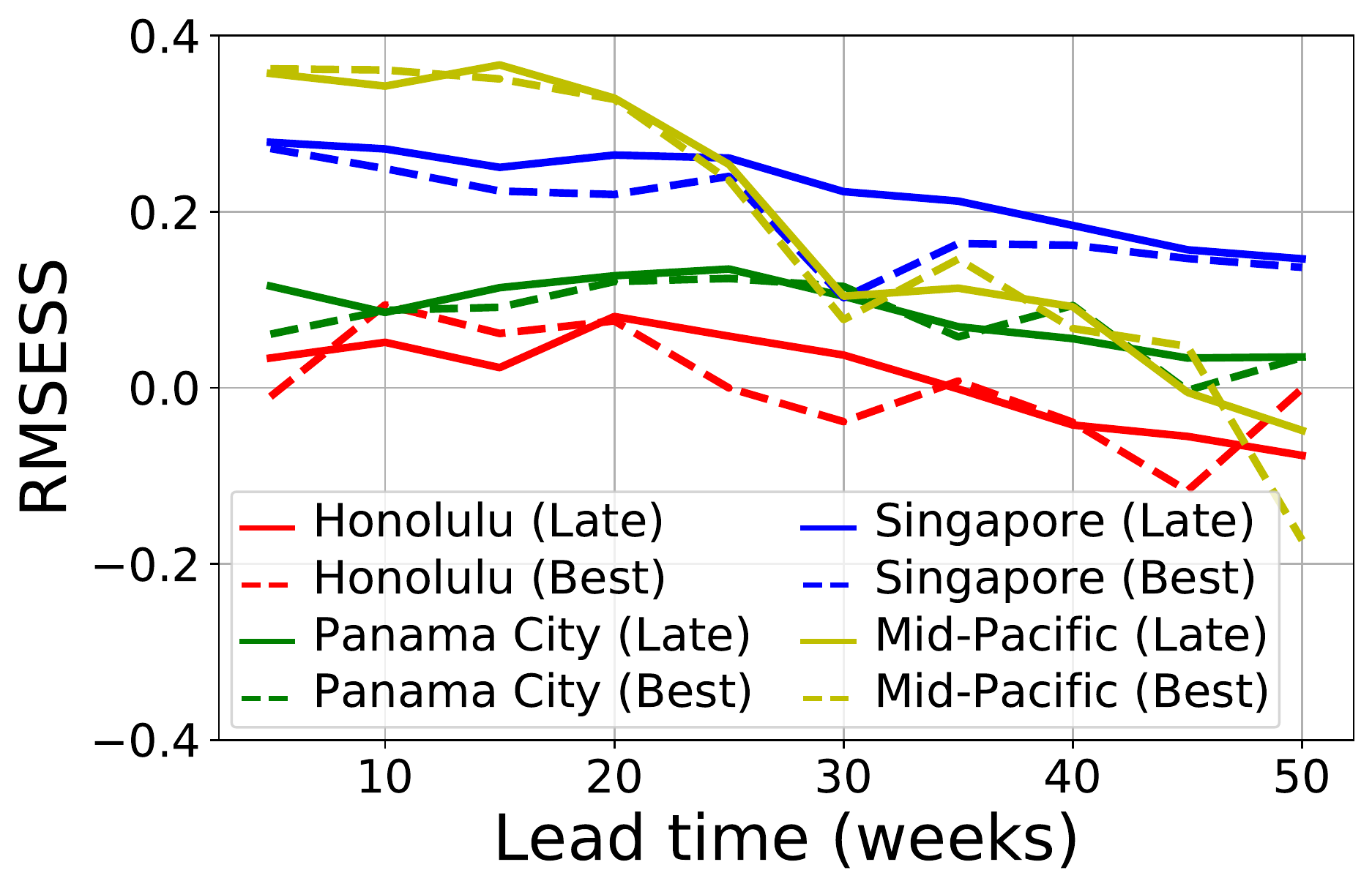}
    \end{minipage}
    \hspace{2em}
    \begin{minipage}[t]{0.40\linewidth}
        \scriptsize
        \centering
        \includegraphics[width=1\linewidth]{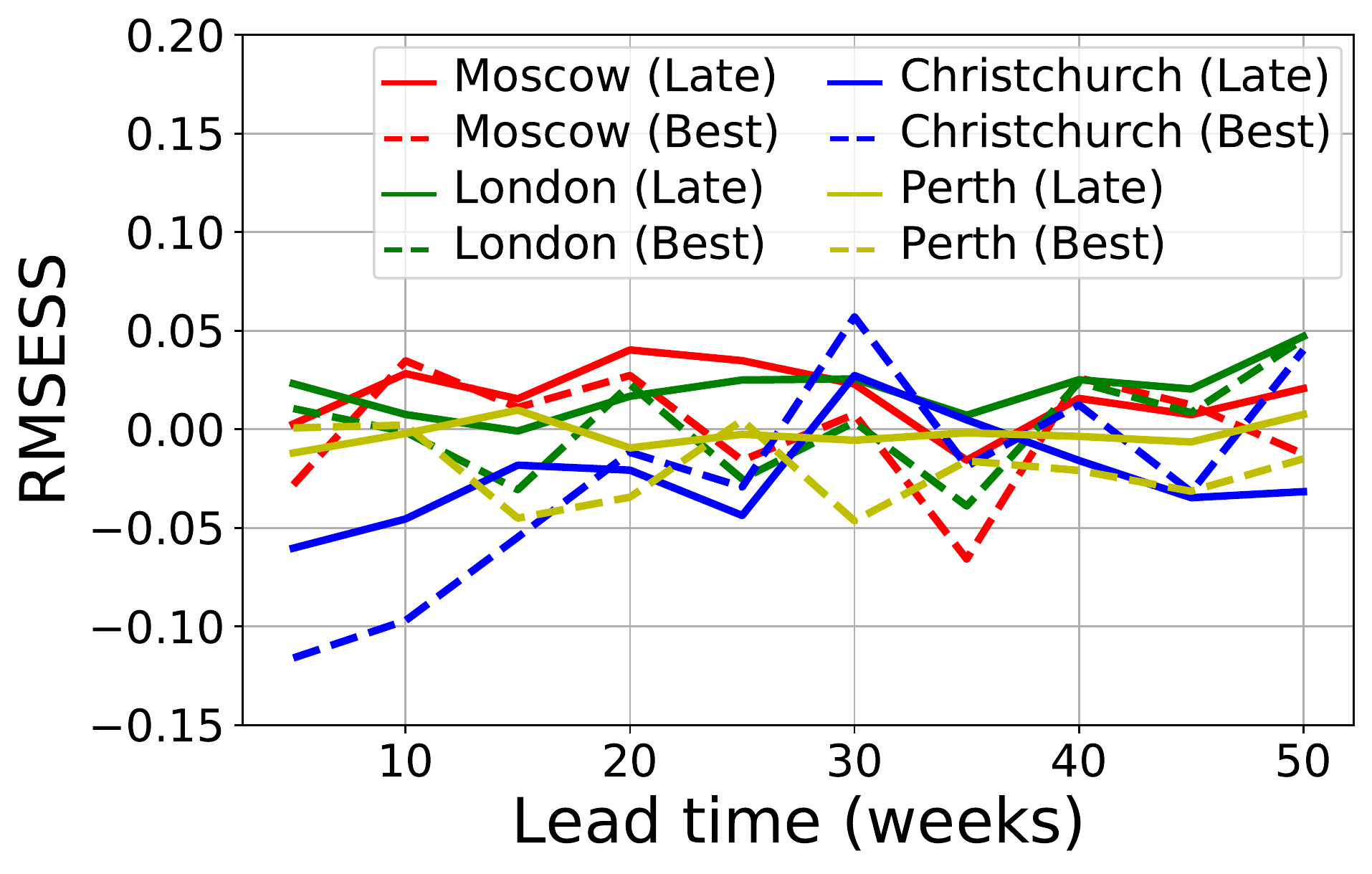}
    \end{minipage}
    \caption{Performance of the late fusion and best model frameworks with 20 models per lead time.}
    \label{fig:20models}
\end{figure}

\subsection{Results}

Fig. \ref{fig:predictions} (left) shows examples of forecasts on the testing data before applying the frameworks. In Singapore, with a lead time of five weeks, the forecasts closely followed the ground truth and outperformed the climate normals especially in 2016. In fact, 2016 was the hottest year on record \cite{url:noaa2021:hottest} and the proposed model was able to forecast this anomalous event. However, as expected, the accuracy decreased with the increase of the lead time. In London, both the forecasts and the climate normals were very similar to the ground truth regardless of the lead time, probably because of the larger range in temperature.

In the RMSESS plot of Singapore in Fig. \ref{fig:predictions} (right), the mostly positive scores indicate that the forecasts outperformed the climate normals, though the scores decreased when the lead time increased. In London, the forecasts and climate normals were very similar, and the discrepancies among models were less obvious. Both plots show that although identical hyperparameters were used in training, the models performed differently especially with large lead times. By combining these models, the late fusion framework outperformed the best model framework and had the best overall results.

Fig. \ref{fig:vs_num_models} shows comparison between the late fusion and the best model frameworks. The late fusion framework outperformed the best model framework in general. When the number of models per lead time increased, the late fusion framework improved smoothly in most locations and gradually converged with around 16 models. In contrast, the best model framework performed less well and may not benefit from a larger number of models. This is because the late fusion framework systematically reduced the expected errors from all models, while the best model framework only chose a single model that had the overall minimal RMSE on the validation data. Fig. \ref{fig:20models} compares the two frameworks with 20 models per lead time. The late fusion framework outperformed the best model framework at most locations.

\section{Conclusion}

The results show that the models trained by the proposed architecture and training strategy can forecast large deviations from climate normals that attribute to climate change. Nevertheless, the models trained with identical hyperparameters may perform differently especially with large lead times. Using the late fusion approach, predictions from different models are combined systematically to provide forecasts with reduced expected errors, and the results can be better than using a single model with the least validation error. As late fusion also improves forecasts with large lead times which associate with large model uncertainties, it is valuable for long-range climate forecasting.

\bibliographystyle{plain}
\bibliography{Ref}

\end{document}